\documentclass[letterpaper]{article} 
\usepackage{aaai25}  
\usepackage{times}  
\usepackage{helvet}  
\usepackage{courier}  
\usepackage[hyphens]{url}  
\usepackage{graphicx} 
\usepackage{natbib}  
\usepackage{caption} 
\usepackage{algorithm}
\usepackage{algorithmic}
\usepackage{subcaption}
\usepackage{soul}
\usepackage{newfloat}
\usepackage{listings}
\DeclareCaptionStyle{ruled}{labelfont=normalfont,labelsep=colon,strut=off} 
\floatstyle{ruled}
\newfloat{listing}{tb}{lst}{}
\floatname{listing}{Listing}
\usepackage{booktabs}
\usepackage{multirow}
\title{Improving Representation Learning of Complex\\Critical Care Data with ICU-BERT}

\author {
    Ricardo Santos\textsuperscript{\rm 1,\rm 2},
    André V. Carreiro\textsuperscript{\rm 1},
    Xi Peng\textsuperscript{\rm 3},
    Hugo Gamboa\textsuperscript{\rm 1,\rm 2},
    Holger Fröhlich\textsuperscript{\rm 4,5}
}
\affiliations {
    \textsuperscript{\rm 1}Fraunhofer AICOS, Portugal\\
    \textsuperscript{\rm 2}LibPhys, NOVA School of Science and Technology, Portugal\\
    \textsuperscript{\rm 3}University of Delaware, DE, USA\\
    \textsuperscript{\rm 4}Fraunhofer SCAI, Department of Bioinformatics, 53757 Sankt Augustin Germany\\
    \textsuperscript{\rm 5}University of Bonn, Bonn-Aachen International Center for IT (b-it), Bonn, 53115 Germany\\
    \{ricardo.santos, andre.carreiro, hugo.gamboa\}@aicos.fraunhofer.pt, holger.froehlich@scai.fraunhofer.de, xipeng@udel.edu
}

\usepackage{bibentry}

\begin{document}

\maketitle

\begin{abstract}

The multivariate, asynchronous nature of real-world clinical data, such as that generated in Intensive Care Units (ICUs), challenges traditional AI-based decision-support systems. These often assume data regularity and feature independence and frequently rely on limited data scopes and manual feature engineering. 
The potential of generative AI technologies has not yet been fully exploited to analyze clinical data.
We introduce ICU-BERT, a transformer-based model pre-trained on the MIMIC-IV database using a multi-task scheme to learn robust representations of complex ICU data with minimal preprocessing. 
ICU-BERT employs a multi-token input strategy, incorporating dense embeddings from a biomedical Large Language Model to learn a generalizable representation of complex and multivariate ICU data.
With an initial evaluation of five tasks and four additional ICU datasets, ICU-BERT results indicate that ICU-BERT either compares to or surpasses current performance benchmarks by leveraging fine-tuning.
By integrating structured and unstructured data, ICU-BERT advances the use of foundational models in medical informatics, offering an adaptable solution for clinical decision support across diverse applications.

\end{abstract}

\section{Introduction}

Generative AI holds great potential to revolutionize healthcare applications by enabling the processing and integration of vast amounts of diverse medical data. These models excel at synthesizing information from structured data like laboratory results, semi-structured machine outputs, and unstructured formats such as clinical notes and imaging reports. 

The Intensive Care Unit (ICU) is a setting where such innovations can be transformative. ICUs generate an immense volume of high-fidelity data from monitoring equipment, electronic health records (EHRs), and clinician notes. These data streams are multivariate, asynchronous, and often multimodal, combining categorical, ordinal, and continuous variables collected at different resolutions. This complexity poses significant challenges to traditional Decision Support Systems (DSS), which rely on assumptions about data regularity and completeness \cite{johnson_mimic-iv_2023}.

Large ICU databases like MIMIC-IV \cite{johnson_mimic-iv_2023, goldberger_physiobank_2000} have enabled significant research progress, but existing DSS often fall short of real-world deployment \cite{eini-porat_tell_2022}. Traditional approaches often employ interpolation, imputation, or resampling to handle missing and irregular data, which can distort natural patterns and relationships.

More sophisticated models have emerged since the Transformer \cite{vaswani_attention_2017}, producing embeddings from sparse longitudinal data, such as BEHRT \cite{li_behrt_2020}, Med-BERT \cite{rasmy_med-bert_2021}, ExMed-BERT \cite{lentzen_exmed-bert_2023}, STraTS \cite{tipirneni_strats_2022} or DuETT \cite{labach_duett_2023}.
These models often rely on narrow data scopes, such as International Classification of Diseases (ICD) codes \cite{hirsch_icd-10_2016} or limited clinical variables. 

In response to these challenges, we introduce ICU-BERT, a generalizable Transformer model based on BERT \cite{devlin_bert_2019} designed to handle the sparse, multivariate, irregular sampling nature of ICU data. ICU-BERT processes medical data sequentially, where each token represents a single entry in a medical record, towards enhancing data representation and predictive performance.

ICU-BERT introduces a novel multi-token input strategy that more effectively captures the intricate details of medical data streams. 
We propose a quadruplet representation of medical registries and a multi-layer embedding structure to improve clinical context interpretation.
Furthermore, pre-trained textual embeddings provide a sophisticated initial token representation of medical concepts, facilitating robust generalization across varied clinical settings and data structures.
To learn complex relationships, ICU-BERT employs a novel pre-training masking task coupled with a multi-task learning loss. 
We pre-trained ICU-BERT on the MIMIC-IV database, providing it with 
adaptable and advanced data representations between clinical variables and their values.

In a preliminary evaluation experiment, ICU-BERT demonstrated robust generalization in multiple real-world challenges. The model was fine-tuned in five tasks from DuETT \cite{labach_duett_2023} and the Yet Another ICU Benchmark (YAIB) framework \cite{van_de_water_yet_2024}, encompassing a range from classification to regression and from single-shot evaluations to continuous monitoring.
Fine-tuning was performed on MIMIC-IV and four YAIB-processed data sets. ICU-BERT benefits from its rich input structure and compares favorably with existing models, underscoring its potential to revolutionize DSS in critical care.

ICU-BERT contributes to the field by i) introducing a multi-token input strategy, a novel masking technique, and a multi-task pre-training scheme; ii) enhancing generalization and expanding usable clinical variables through robust pre-trained textual embeddings; and iii) outperforming models with limited features and simple data structures in real-world tasks, surpassing benchmarks in some tasks.

\section{Related Work}

Using Machine Learning (ML) to process extensive health data has been a long-standing effort, particularly in ICU settings where data is collected from various devices at minute or second intervals \cite{wiens_machine_2018}. Traditional DSS require extensive preprocessing and feature extraction \cite{kong_using_2020}. Recurrent models capture temporal relationships but struggle with sparse, asynchronous ICU data due to fixed temporal resolutions and reliance on predefined variable matrices \cite{ge_interpretable_2018}.

Advancements in Transformers improved the modeling of longitudinal medical data. BEHRT \cite{li_behrt_2020} processes diagnosis codes as tokens within a limited vocabulary, modeling hospital visits as sentences with age encoded in position embeddings and pre-training through Masked Language Modeling (MLM). Building on this, Med-BERT \cite{rasmy_med-bert_2021} and ExMed-BERT \cite{lentzen_exmed-bert_2023} extended the approach by incorporating continuous data and adopting late fusion techniques. Models like Rare-BERT \cite{prakash_rarebert_2021} and life2vec \cite{savcisens_life2vec_2023} further expanded vocabularies to cover broader data types, while ExBEHRT \cite{rupp_exbehrt_2023} and BRLTM \cite{meng_brltm_2021} integrated demographic and lab data through additional embedding layers.

Hierarchical approaches such as Hi-BEHRT \cite{li_hi-behrt_2021} and graph-based methods like GT-BEHRT \cite{poulain_gt-behrt_2023} addressed challenges with large input sequences by summarizing sets of visits before integrating them into patient profiles. STraTS \cite{tipirneni_strats_2022} proposed a triplet embedding scheme for features, values, and time, while DuETT \cite{labach_duett_2023} used dual attention to capture time- and event-based representations.

A key challenge remains the lack of standardized benchmarks to compare. YAIB \cite{van_de_water_yet_2024} provides a reproducible framework for ICU cohort generation and task evaluation, while EHRSHOT \cite{wornow_ehrshot_2023} enables model fine-tuning across contexts.

ICU-BERT advances these efforts by capturing the complexity of ICU data through pre-trained embeddings and modeling temporal information at the embedding level. Unlike earlier approaches, it avoids narrow vocabularies, enabling generalization across datasets and achieving competitive performance on YAIB and DuETT tasks.

\section{Methods}

\begin{figure*}[ht!]
\centering
\includegraphics[width=.9\linewidth]{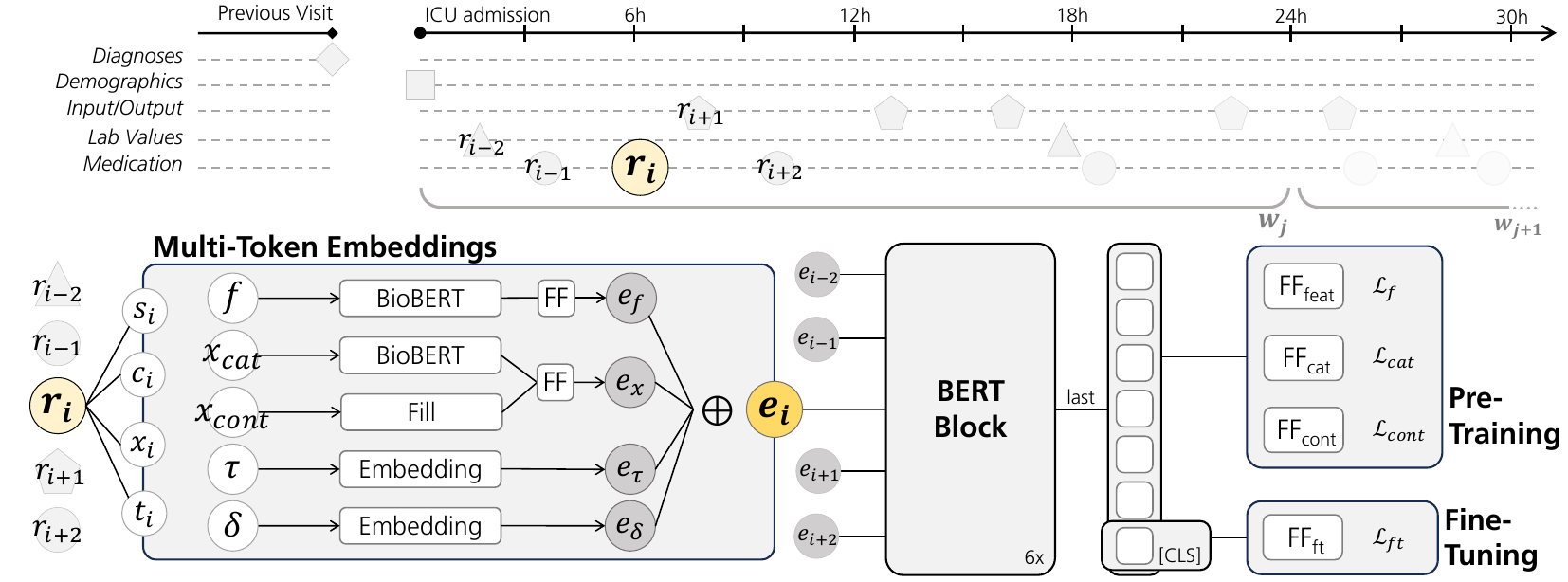}
\caption{ICU-BERT scheme. Complex, multivariate, and sparse medical registries $r_i$ are processed by a multi-token embedding structure that combines embeddings from feature names $f$, categorical or numerical values $x$, timestamps $\tau$, and durations $\delta$. Pre-trained embeddings enhance the representations of features and categorical values, and a novel pre-training multi-task loss optimizes the simultaneous reconstruction of both features and values.}
\label{fig:scheme}
\end{figure*}

ICU-BERT employs a bidirectional Transformer architecture \cite{devlin_bert_2019}, optimized for high-dimensional, sparse, and multivariate ICU data with varying time resolutions and sampling rates. The model uses a multi-token input strategy, representing each medical registry as a quadruplet of clinical variable, value, recording time, and duration.

ICU-BERT integrates pre-trained text embeddings from BioBERT \cite{lee_biobert_2020}, a biomedical domain-specific Large Language Model, to provide robust representations of medical concepts and improve generalization. Unlike traditional vocabularies, which often fail to reflect nuanced clinical relationships, such as differences in blood pressure definitions across contexts, BioBERT embeddings preserve semantic meaning without extensive harmonization.

To optimize learning, ICU-BERT uses a novel masking task and multi-task learning loss during pre-training, enabling effective reconstruction of clinical variables and values. Pre-training on the entire MIMIC-IV database allows the model to develop sophisticated representations without extensive pre-processing, including feature engineering or semantic harmonization.

ICU-BERT enhances the interpretation of complex ICU data, making it highly adaptable for real-world clinical applications. Figure \ref{fig:scheme} provides an overview of the architecture.

\subsection{Data Representation}

In ICU-BERT, we conceptualize medical data for all $P$ patients in a database. $V_p$ integrates all ICU stays of patient $p \in \{1, 2, ..., P\}$. Each ICU stay $v \in V_p$ contains a set $R_v$, which includes all registries $r$ recorded during that stay. Each registry $r_i, i \in \{1, 2, ..., N\}$, with $N$ the total number of observations, is characterized by four attributes: the data source $s$, the clinical variable $c$, the recorded value $x$, and the associated timestamp or interval $t$. Mathematically, we can represent each registry as $r_i=(s_i,c_i,x_i,t_i)_{i=1}^N$.

To capture temporal dynamics and granular details, each ICU stay $v$ is divided into non-overlapping 24-hour windows $w_j, j \in \{1, 2, ..., K_v\}$, where $K_v$ is the total number of windows for $v$. Each window $w_j$ includes all data entries $r_i$ with timestamps $t_i$ within the window's start and end boundaries, along with static demographic data, admission ward, and documented prior diagnoses.

Each registry $r_i$ within a window $w_j$ is converted into a quadruplet representation $q_i = (f(s, c), x, \tau(t), \delta(t))$, which serves as a token. Here, $f(s, c)$ represents the feature derived from the textual aggregation of source $s$ and clinical variable $c$, $\tau(t)$ encodes the relative time since the start of $w_j$, and $\delta(t)$ captures the duration associated with $t$, defaulting to $0$ for discrete events. A classification token $[\mathrm{CLS}]$ and padding tokens $[\mathrm{PAD}]$ are added, maintaining the same quadruplet structure.

\subsection{Multi-Token Embeddings}

To enhance the representational power of ICU-BERT and the ability to interpret and process medical terminology and context effectively, each $f(s, c)$ within the quadruplets $q_i$ is enriched with BioBERT pre-trained embeddings, extracted from the token $[\mathrm{CLS}]$ from the last hidden state, such as:

\begin{equation}
    PreE_{f} = \mathrm{BioBERT}(f(s, c))[\mathrm{CLS}]
\end{equation}

For each quadruplet $q_i$, a Boolean mask $m_i$ is defined to denote whether the corresponding value
$x$ is continuous ($m_i = 1$). BioBERT embeddings are extracted for categorical values, while a repetition of the continuous values matches the $768$ dimension of the pre-trained embedding:

\begin{equation}
    PreE_{x} = \left\{
    \begin{array}{ll}
          \mathrm{fill}(x), & \mathrm{if} \quad m_i = 1 \\
          \mathrm{BioBERT}(x)[\mathrm{CLS}], & \mathrm{if} \quad m_i = 0
    \end{array} 
    \right.
\end{equation}

Following the initial extraction of pre-trained embeddings, ICU-BERT employs a refined embedding layer to adapt these embeddings for ICU data analysis. Pre-trained embeddings of features $f$ and values $v$ undergo a dense transformation to $d$, the input dimension of the BERT block:

\begin{equation}
    e_f = W_f\cdot PreE_f + b_f
\end{equation}

\begin{equation}
    e_x = W_x \cdot PreE_x + b_x
\end{equation}

$W_f$ and $W_x$ represent the weight matrices for features and values, respectively, and $b_f$ and $b_x$ are the bias terms. 

For encoding temporal information such as timestamps and durations, ICU-BERT employs an embedding layer that maps each discrete time-related input, in minutes, to a high-dimensional continuous vector. Specifically, the embedding for time $\tau(t)$ and duration $\delta(t)$ are obtained by indexing into a pre-defined embedding matrix of input size 1,440 (minute-level in 24 hours), which transforms into $d$, represented as:

\begin{equation}
    e_{\tau} = Embedding({\tau}(t))
\end{equation}

\begin{equation}
    e_{\delta} = Embedding({\delta}(t))
\end{equation}

The embeddings $e_f$, $e_x$, $e_{\tau}$ and $e_{\delta}$ are summed to produce the composite embedding of $r_i$:

\begin{equation}
    e_i = e_f + e_x + e_{\tau} + e_{\delta}
\end{equation}

The resulting vector is subjected to dropout followed by layer normalization to ensure that the embeddings have a consistent scale and distribution. $e_i$ thus encapsulates a detailed and nuanced representation of static and dynamic patient data, enabling precise and context-aware predictions.

\subsection{BERT Configuration}

ICU-BERT uses a BERT architecture \cite{devlin_bert_2019}, adopting the original configuration of six transformer layers, each comprising multi-head self-attention mechanisms and fully connected networks. By applying this well-established architecture to ICU data, ICU-BERT benefits from BERT's powerful feature extraction capabilities, making it effective for interpreting complex medical data.

\subsection{Pre-training Approach}

Pre-training ICU-BERT is critical for adapting BERT’s language capabilities to ICU data.
ICU-BERT introduces a novel pre-training strategy, Masked Language-Value Modelling (MLVM), that selects 15\% of the quadruplets for masking.
To enhance the model's ability to reconstruct clinical features and their values, among the selected quadruplets, 50\% have both their feature name and value masked, 25\% have only the value masked, and 25\% have only the feature name masked. Similar to MLM, masked elements are replaced with a $[\mathrm{MASK}]$ token, substituted with a random token, or left unchanged in an 80\%-10\%-10\% ratio.

MLVM reconstruction uses three output heads with separate vocabularies. The feature name vocabulary comprises all possible feature names, with size $\mathcal{F}$, while the value vocabulary includes all possible categorical values, with size $\mathcal{V}$. 
The feature classification head predicts the identity of masked features using an output dimension equal to the size of the feature name vocabulary. 
Instead, the categorical value classification head predicts the identity of masked categorical values.
Lastly, the continuous value regression head is designed to predict continuous values and outputs a single scalar directly.

\subsection{Multi-Task Loss}

ICU-BERT employs distinct loss functions in a multi-task framing to optimally train each of its classification heads, reflecting the varied nature of the medical data.
Cross-entropy loss is used to predict masked features and categorical values. However, the loss of categorical values must account for the continuous mask, $m$:

\begin{equation}
    \mathcal{L}_f = -\frac{1}{K} \sum_{k=1}^{K} \sum_{c=1}^{\mathcal{F}} y_{k, c} log(\hat{y}_{k, c})
\end{equation}

\begin{equation}
    \mathcal{L}_{cat} = -\frac{1}{K - \sum_{k=1}^{K} m_k} \sum_{k=1}^{K} (1 - m_k) \sum_{c=1}^{\mathcal{V}} y_{k, c} log(\hat{y}_{k, c})
\end{equation}

$K$ is the total number of tokens in the batch. The reconstruction loss of continuous values uses the Mean Absolute Error (MAE):

\begin{equation}
    \mathcal{L}_{cont} = -\frac{1}{\sum_{k=1}^{K} m_k} \sum_{k=1}^{K} m_k | y_k - \hat{y}_k |
\end{equation}

Given the losses for features $\mathcal{L}_{f}$, categorical $\mathcal{L}_{cat}$, and continuous values $\mathcal{L}_{cont}$, the multi-task loss $\mathcal{L}_{total}$ becomes:

\begin{equation}
    \mathcal{L}_{total} = \mathcal{L}_{f} + \beta \times \frac{\mathcal{L}_{cat} \times N_{cat} + \alpha  \times\mathcal{L}_{cont} \times N_{cont} }{N_{cat} + N_{cont}}
\end{equation}

$\alpha$ and $\beta$ are parameterized weights to balance the contribution of value losses, while $N_{cont} = \sum_{k=1}^{K} m_k$ and $N_{cat} = K - N_{cont}$.
This strategy not only challenges the model to predict the missing parts but also encourages it to learn robust associations between features and values.

\subsection{Fine-Tuning}

While the rich reconstructions from pre-training do not have direct clinical value, ICU-BERT can be fine-tuned to specific tasks. The original classification head needs to be replaced with an appropriate one, along with a suitable loss function. Outputs for fine-tuning are derived from the $[\mathrm{CLS}]$ token of the final transformer layer, which encapsulates the entire input sequence's contextual information. This ensures that the predictions are based on comprehensive patient data, improving the applicability of ICU-BERT in clinical settings.

\section{Experiments}

Despite the limited availability of benchmarks to evaluate ICU-BERT \cite{van_de_water_yet_2024}, we conducted a set of experiments to assess its robustness and applicability. 

ICU-BERT was pre-trained on the MIMIC-IV v2.2 dataset \cite{johnson_mimic-iv_2023}, with the relational database structured into a hierarchical schema of timestamped entries for each clinical variable and data source. Data from the \textit{hosp} and \textit{icu} tables were used, partitioned into train, validation, and test sets in a 70\%-15\%-15\% split, following DuETT's \cite{labach_duett_2023} methodology.

For fine-tuning, ICU-BERT was adapted to a diverse set of clinical tasks, including two from DuETT and three from YAIB \cite{van_de_water_yet_2024}, chosen to demonstrate the model's versatility. To assess generalization, we conducted zero-shot external evaluations and fine-tuned the model across four additional datasets. A 5-fold cross-validation (CV) approach was used, resampling train and validation sets for robust performance assessment.

\subsection{Datasets and Real-World Tasks}

We extracted structured datasets from HiRID v1.1.1 \cite{faltys_hirid_2021}, eICU v2.0 \cite{pollard_eicu_2018}, MIMIC-III v1.4 \cite{johnson_mimic-iii_2016}, and again MIMIC-IV v2.2 but with a limited feature set, using YAIB standards. These datasets are configured to include 52 features that encompass static and time series data, such as vital signs, laboratory results, and input/output. The \textit{ricu} R package \cite{bennett_ricu_2022} is used within the framework to homogeneously process ICU databases, from which specific cohorts are extracted and structured in a tabular format.

The fine-tuning process was carried out in five distinct classification and regression tasks to assess the model's predictive power and applicability to real-world clinical scenarios. Cohorts and targets for YAIB datasets are predefined, and the same rationale was applied in creating MIMIC-IV's fine-tuning samples.

\subsubsection{Hospital Mortality}
From DuETT, this task is predicted as a binary outcome based on the first 48 hours of data, using the remaining stay until death or discharge as the target.

\subsubsection{Phenotyping}
A multi-label classification task is defined from DuETT that consists of phenotyping 25 diseases, evaluated at discharge, based on the first 24 hours of ICU stay.

\subsubsection{ICU Mortality}
From YAIB, predicting ICU mortality within the first 24 hours is adopted as a binary classification.

\subsubsection{Kidney Function}
This regression task involves predicting the median creatinine levels on day two after ICU admission, based on the first 24 hours.

\subsubsection{AKI Onset}
The onset of acute kidney injury (AKI) is predicted using the KDIGO 2012 criteria \cite{kdigo_section_2012}. While YAIB uses a recurrent many-to-many approach predicting hourly onset from all prior data, ICU-BERT employs a rolling 24-hour window with a 6-hour step to accommodate sequence length limits.

Table \ref{tab:datasets} shows the summary of learning samples in each dataset from the cohort extraction for each task, displaying the high imbalance common in ICU-related tasks.

\begin{table*}
\centering
\begin{tabular}{lccccc}
\toprule
 & MIMIC-IV & YAIB: MIMIC-IV & YAIB: MIMIC-III & YAIB: eICU & YAIB: HiRID \\ 
\midrule
Patients & 50,920 & 50,920 & 46,476 & 160,816 & N.A. \\ 
ICU Stays & 73,181 & 73,181 & 61,532 & 182,774 & 32,338 \\ 
\midrule
ICU Mortality & 49,833 (9.1\%) & 49,523 (7.2\%) & 38,128 (8.1\%) & 113,381 (5.5\%) & 12,859 (8.5\%) \\ 
Kidney Function  & 33,677 (1.0) & 33,949 (1.0) & 27,481 (1.0) & 69,116 (1.0) & 7,499 (0.9) \\ 
Onset AKI &  314,715 (13.1\%) & 478,582 (11.7\%) & 388,951 (8.7\%) & 1,177,049 (11.3\%) & 202,681 (7.8\%) \\ 
Phenotyping &  48,998 (18.6\%) & \multicolumn{4}{c}{\multirow{2}{*}{N.A.}}  \\ 
Hospital Mortality  & 35,131 (14.2\%) & \multicolumn{4}{c}{} \\ 
\bottomrule
\end{tabular}
\caption{Summary of total samples in included datasets and fine-tuning tasks. Classification tasks include the positive rate (incidence) of the event, while the regression task (Kidney Function) includes the median value in mg/dL.} 
\label{tab:datasets}
\end{table*}

\subsection{Implementation Details}

ICU-BERT was developed using PyTorch 2.3.0 and the HuggingFace Transformers 4.40.2 \cite{wolf_huggingfaces_2020}, on Linux with CUDA 12.2. Pre-training was performed on eight 40GB NVIDIA Tesla A100 GPUs (1.9 TB VRAM) using SMX4 protocol, while fine-tuning utilized NVIDIA Tesla A10 GPUs (205.4 GB VRAM). 

The model follows Med-BERT specifications \cite{rasmy_med-bert_2021}, with 6 encoder layers, 768 hidden size, 6 attention heads, a 64-dimensional dense filter, and a maximum sequence length of 512 tokens.

ICU-BERT was pre-trained on the MIMIC-IV training set ($\mathcal{F}$ = 45,825; $\mathcal{V}$ = 1,026) for 102 epochs over 10.4 days. Optimization used AdamW with a 5e-5 learning rate, 0.1 dropout, no weight decay, and a linear scheduler with a 40-epoch warmup. Hyperparameters $\alpha$ and $\beta$ were set to 3 and 1 after tuning on 10\% of the training data.

Fine-tuning ran for up to 50 epochs with early stopping after 10 epochs based on validation loss. Hyperparameter tuning via Optuna \cite{akiba_optuna_2019} targeted the ICU mortality task over 25 epochs. The best setup used the $[\mathrm{CLS}]$ token from the last layer, with the last five layers unfrozen. The AdamW optimizer had a 1e-3 learning rate, 0.5 final dropout, and a linear scheduler without warmup.

\section{Results}

The performance of classification tasks was measured using the Area Under the Receiver Operating Characteristic (AUROC) and the Area Under the Precision-Recall Curve (AUPRC). For regression, MAE was employed. 

Table \ref{tab:yaib_metrics} shows that ICU-BERT achieved an AUROC of $88.9\pm0.3\%$ and an AUPRC of $48.5\pm0.7\%$ for ICU mortality. The low AUPRC reflects the severe class imbalance (9.1\% incidence) despite class weight adjustments. A similar trend was observed across other tasks.

\begin{table*}[htb!]
\centering
\begin{tabular}{lccccc}
\toprule
\multicolumn{1}{c}{\multirow{2}{*}{Model}} & \multicolumn{2}{c}{Mortality ICU} & \multicolumn{2}{c}{Onset AKI} & Kidney Function\\
\cmidrule(l){2-6}
                           & AUROC $\uparrow$ & AUPRC $\uparrow$ & AUROC $\uparrow$ & AUPRC $\uparrow$ & MAE $\downarrow$ \\
\midrule
ICU-BERT & \textbf{0.889 $\pm$ 0.003} & \textbf{0.485 $\pm$ 0.007} & 0.824 $\pm$ 0.008 & 0.468 $\pm$ 0.017 & 0.248 $\pm$ 0.008 \\
\midrule
Logistic Regression       & 0.861 $\pm$ 0.001 & 0.397 $\pm$ 0.006 & 0.771 $\pm$ 0.002 & 0.377 $\pm$ 0.003 & N.A. \\
LGBM    & 0.877 $\pm$ 0.002 & 0.442 $\pm$ 0.007 & 0.838 $\pm$ 0.001 & 0.533 $\pm$ 0.002 & \textbf{0.24 $\pm$ 0.00} \\
GRU    & 0.876 $\pm$ 0.001 & 0.428 $\pm$ 0.003 & \textbf{0.907 $\pm$ 0.001} & \textbf{0.696 $\pm$ 0.002} & 0.30 $\pm$ 0.01 \\
LSTM     & 0.867 $\pm$ 0.004 & 0.410 $\pm$ 0.007 & 0.897 $\pm$ 0.001 & 0.665 $\pm$ 0.002 & 0.28 $\pm$ 0.01 \\
TCN       & 0.871 $\pm$ 0.003 & 0.414 $\pm$ 0.008 & 0.898 $\pm$ 0.001 & 0.668 $\pm$ 0.002 & 0.28 $\pm$ 0.01 \\
Transformer      & 0.869 $\pm$ 0.003 & 0.422 $\pm$ 0.003 & 0.896 $\pm$ 0.001 & 0.656 $\pm$ 0.002 & 0.32 $\pm$ 0.01 \\
Elastic Net       & \multicolumn{4}{c}{N.A.}                             & 0.25 $\pm$ 0.00 \\
\bottomrule
\end{tabular}
\caption{Performance of ICU-BERT in the YAIB tasks after fine-tuning on MIMIC-IV. The remaining models are baselines from \cite{van_de_water_yet_2024}. The best result is emboldened. N.A. indicates that the model was not trained in the task.}
\label{tab:yaib_metrics}
\end{table*}

\begin{figure}[htb!]
\centering
\includegraphics[width=\linewidth]{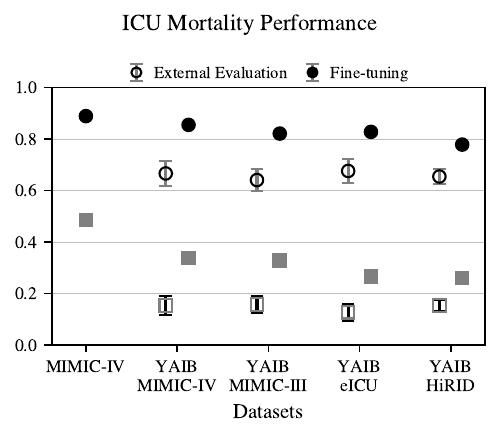}
\caption{Results for ICU mortality as mean and standard deviation over 5-fold CV, on external evaluation and fine-tuning, in MIMIC-IV and the YAIB-processed datasets.}
\label{fig:mortality_plot}
\end{figure}

External zero-shot evaluations on YAIB datasets revealed a drop in AUROC 64.1\%-67.6\% and AUPRC to 12.7\%-15.8\% in ICU mortality, as seen in Figure \ref{fig:mortality_plot}. Similar declines were noted in the AKI onset and kidney function tasks, particularly when applying the model to distinct patient populations in new datasets such as eICU and HiRID. However, fine-tuning significantly improved results, highlighting ICU-BERT's potential as a foundational ICU model.

ICU-BERT’s multi-token embedding strategy effectively captures the semantic and numerical complexity of clinical variables, bypassing the need for pre-processing. This allows the model to utilize all available data and enhance representation learning. A comparison using two versions of MIMIC-IV revealed a performance drop when limited to 52 features, underscoring the importance of a rich feature set.

\begin{table*}
\centering
\begin{tabular}{lcccc}
\toprule
\multicolumn{1}{c}{\multirow{2}{*}{Model}} & \multicolumn{2}{c}{Hospital Mortality} & \multicolumn{2}{c}{Phenotyping}\\
\cmidrule(l){2-5}
                           & AUROC $\uparrow$ & AUPRC $\uparrow$ & AUROC $\uparrow$ & AUPRC $\uparrow$ \\
\midrule
ICU-BERT  & 0.875 $\pm$ 0.007 & 0.558 $\pm$ 0.014 & \textbf{0.844 $\pm$ 0.001} & 0.573 $\pm$ 0.003 \\
\midrule
DuETT     & \textbf{0.912 $\pm$ 0.020} & \textbf{0.627 $\pm$ 0.002} & 0.838 $\pm$ 0.001 & \textbf{0.604 $\pm$ 0.003} \\
STraTS    & 0.882 $\pm$ 0.004 & 0.552 $\pm$ 0.013 & 0.820 $\pm$ 0.001 & 0.565 $\pm$ 0.006 \\
Raindrop \citep{zhang_graph-guided_2022} & 0.878 $\pm$ 0.001 & 0.546 $\pm$ 0.002 & 0.824 $\pm$ 0.001 & 0.577 $\pm$ 0.002 \\
mTAND \citep{shukla_mtan_2021}    & 0.864 $\pm$ 0.002 & 0.540 $\pm$ 0.007 & 0.812 $\pm$ 0.001 & 0.553 $\pm$ 0.007 \\
LSTM      & 0.881 $\pm$ 0.001 & 0.522 $\pm$ 0.006 & 0.756 $\pm$ 0.002 & 0.447 $\pm$ 0.002 \\
XGBoost \citep{chen_xgboost_2016}  & 0.886 $\pm$ 0.002 & 0.593 $\pm$ 0.004 & 0.829 $\pm$ 0.001 & 0.589 $\pm$ 0.009 \\

\bottomrule
\end{tabular}
\caption{Performance of ICU-BERT model in the ICU tasks replicated from DuETT after fine-tuning on the MIMIC-IV dataset. The remaining models are baselines published by \cite{labach_duett_2023}. The best result is emboldened.}
\label{tab:duett_metrics}
\end{table*}

As shown in Tables \ref{tab:yaib_metrics} and \ref{tab:duett_metrics}, ICU-BERT set new benchmarks for ICU mortality and closely matched baselines in kidney function and phenotyping. However, it underperformed in predicting AKI onset, a task requiring continuous modeling. This limitation arises from BERT’s 512-token input constraint, which we addressed by pre-training ICU-BERT with 24-hour windows. While recurrent models performed better in tasks requiring temporal continuity, they often failed to fully exploit detailed patient data.

\section{Discussion}

A foundational model must leverage all available patient data over time. Future improvements could include hierarchical strategies, such as additional recurrent layers or advanced transformers like Hi-BERT, Longformer \cite{beltagy_longformer_2020}, or Mamba \cite{gu_mamba_2023}, to handle longer sequences and reduce the segmentation of continuous tasks into discrete windows.

Comparing ICU-BERT to existing models is challenging. While ICU-BERT can process granular ICU data, enabling real-time applications, Med-BERT is limited to processing ICD codes at the visit level. In comparison with DuETT, we replicated partitions and tasks on MIMIC-IV but could not ensure identical cohorts or training conditions. While ICU-BERT does not consistently surpass all models, its adaptability to diverse contexts with minimal processing is a key strength. Its multi-token input strategy and use of pre-trained embeddings capture semantic relationships beyond the constraints of predefined vocabularies.

However, building a foundational ICU model remains challenging. Limited hyperparameter tuning may have constrained performance. Relying solely on the $[\mathrm{CLS}]$ token from the final layer might miss sequence complexities, as pre-training did not include tasks like Next Sequence Prediction. Pre-trained embeddings, while improving feature representation and distinguishing continuous from categorical variables, increase input dimensions and computational costs. Enhancements such as pooling mechanisms, smaller pre-trained models, or advanced continuous value embeddings \cite{gorishniy_embeddings_2023} could address these issues.
Unfreezing specific BERT layers during fine-tuning has shown promising results, though at higher computational costs. 
While ICU-BERT effectively utilizes multivariate data, it has not yet integrated multimodal information, such as clinical notes. Future work should explore pre-training on additional datasets to expand its capabilities.

\section{Conclusion}

ICU-BERT advances representation learning for complex multivariate data, which is crucial for decision support in intensive care. 
The multi-token input strategy and pre-trained textual embeddings allow the model to implicitly capture semantic and numerical relationships across various data types, enhancing the performance of traditional models. This approach overcomes the limitations of previous work that relies on predefined vocabularies, heavily processed structures, and a restricted set of variables, allowing ICU-BERT to efficiently process highly granular and continuous data.

These technical enhancements have implications that extend beyond the ICU settings, offering potential improvements in any domain where complex data streams require accurate and efficient interpretation. Despite these advancements, developing a truly foundational ICU model remains a challenge. Future efforts should aim to expand ICU-BERT's architecture to better accommodate additional data modalities, such as unstructured clinical notes, and handle longer sequence inputs without the need for extensive pre-processing, potentially integrating more efficient architectures to enhance its applicability and performance.

\section{Acknowledgments}

This work was supported by European funds through the Recovery and Resilience Plan via the project ”Center for Responsible AI,” with identification number C645008882-00000055.

\bibliography{references}

\begin{thebibliography}{36}
\providecommand{\natexlab}[1]{#1}

\bibitem[{Akiba et~al.(2019)Akiba, Sano, Yanase, Ohta, and Koyama}]{akiba_optuna_2019}
Akiba, T.; Sano, S.; Yanase, T.; Ohta, T.; and Koyama, M. 2019.
\newblock Optuna: {A} {Next}-generation {Hyperparameter} {Optimization} {Framework}.
\newblock ArXiv:1907.10902 [cs, stat].

\bibitem[{Beltagy, Peters, and Cohan(2020)}]{beltagy_longformer_2020}
Beltagy, I.; Peters, M.~E.; and Cohan, A. 2020.
\newblock Longformer: {The} {Long}-{Document} {Transformer}.
\newblock ArXiv:2004.05150 [cs].

\bibitem[{Bennett et~al.(2022)Bennett, Plečko, Ukor, Meinshausen, and Bühlmann}]{bennett_ricu_2022}
Bennett, N.; Plečko, D.; Ukor, I.-F.; Meinshausen, N.; and Bühlmann, P. 2022.
\newblock ricu: {R}’s interface to intensive care data.
\newblock \emph{GigaScience}, 12: giad041.

\bibitem[{Chen and Guestrin(2016)}]{chen_xgboost_2016}
Chen, T.; and Guestrin, C. 2016.
\newblock {XGBoost}: {A} {Scalable} {Tree} {Boosting} {System}.
\newblock In \emph{Proceedings of the 22nd {ACM} {SIGKDD} {International} {Conference} on {Knowledge} {Discovery} and {Data} {Mining}}, 785--794.
\newblock ArXiv:1603.02754 [cs].

\bibitem[{Devlin et~al.(2019)Devlin, Chang, Lee, and Toutanova}]{devlin_bert_2019}
Devlin, J.; Chang, M.-W.; Lee, K.; and Toutanova, K. 2019.
\newblock {BERT}: {Pre}-training of {Deep} {Bidirectional} {Transformers} for {Language} {Understanding}.
\newblock ArXiv:1810.04805 [cs].

\bibitem[{Eini-Porat et~al.(2022)Eini-Porat, Amir, Eytan, and Shalit}]{eini-porat_tell_2022}
Eini-Porat, B.; Amir, O.; Eytan, D.; and Shalit, U. 2022.
\newblock Tell me something interesting: {Clinical} utility of machine learning prediction models in the {ICU}.
\newblock \emph{Journal of Biomedical Informatics}, 132: 104107.

\bibitem[{Faltys et~al.(2021)Faltys, Zimmermann, Lyu, Hüser, Hyland, Rätsch, and Merz}]{faltys_hirid_2021}
Faltys, M.; Zimmermann, M.; Lyu, X.; Hüser, M.; Hyland, S.; Rätsch, G.; and Merz, T. 2021.
\newblock {HiRID}, a high time-resolution {ICU} dataset (version 1.1.1).

\bibitem[{Ge et~al.(2018)Ge, Huh, Park, Lee, Kim, and Turchin}]{ge_interpretable_2018}
Ge, W.; Huh, J.-W.; Park, Y.~R.; Lee, J.-H.; Kim, Y.-H.; and Turchin, A. 2018.
\newblock An {Interpretable} {ICU} {Mortality} {Prediction} {Model} {Based} on {Logistic} {Regression} and {Recurrent} {Neural} {Networks} with {LSTM} units.
\newblock \emph{AMIA Annual Symposium Proceedings}, 2018: 460--469.

\bibitem[{Goldberger et~al.(2000)Goldberger, Amaral, Glass, Hausdorff, Ivanov, Mark, Mietus, Moody, Peng, and Stanley}]{goldberger_physiobank_2000}
Goldberger, A.~L.; Amaral, L. A.~N.; Glass, L.; Hausdorff, J.~M.; Ivanov, P.~C.; Mark, R.~G.; Mietus, J.~E.; Moody, G.~B.; Peng, C.-K.; and Stanley, H.~E. 2000.
\newblock {PhysioBank}, {PhysioToolkit}, and {PhysioNet}: {Components} of a {New} {Research} {Resource} for {Complex} {Physiologic} {Signals}.
\newblock \emph{Circulation}, 101(23).

\bibitem[{Gorishniy, Rubachev, and Babenko(2023)}]{gorishniy_embeddings_2023}
Gorishniy, Y.; Rubachev, I.; and Babenko, A. 2023.
\newblock On {Embeddings} for {Numerical} {Features} in {Tabular} {Deep} {Learning}.
\newblock ArXiv:2203.05556 [cs].

\bibitem[{Gu and Dao(2023)}]{gu_mamba_2023}
Gu, A.; and Dao, T. 2023.
\newblock Mamba: {Linear}-{Time} {Sequence} {Modeling} with {Selective} {State} {Spaces}.
\newblock ArXiv:2312.00752 [cs].

\bibitem[{Hirsch et~al.(2016)Hirsch, Nicola, McGinty, Liu, Barr, Chittle, and Manchikanti}]{hirsch_icd-10_2016}
Hirsch, J.; Nicola, G.; McGinty, G.; Liu, R.; Barr, R.; Chittle, M.; and Manchikanti, L. 2016.
\newblock {ICD}-10: {History} and {Context}.
\newblock \emph{American Journal of Neuroradiology}, 37(4): 596--599.

\bibitem[{Johnson et~al.(2023)Johnson, Bulgarelli, Shen, Gayles, Shammout, Horng, Pollard, Hao, Moody, Gow, Lehman, Celi, and Mark}]{johnson_mimic-iv_2023}
Johnson, A. E.~W.; Bulgarelli, L.; Shen, L.; Gayles, A.; Shammout, A.; Horng, S.; Pollard, T.~J.; Hao, S.; Moody, B.; Gow, B.; Lehman, L.-w.~H.; Celi, L.~A.; and Mark, R.~G. 2023.
\newblock {MIMIC}-{IV}, a freely accessible electronic health record dataset.
\newblock \emph{Scientific Data}, 10(1): 1.

\bibitem[{Johnson et~al.(2016)Johnson, Pollard, Shen, Lehman, Feng, Ghassemi, Moody, Szolovits, Anthony~Celi, and Mark}]{johnson_mimic-iii_2016}
Johnson, A. E.~W.; Pollard, T.~J.; Shen, L.; Lehman, L.-w.~H.; Feng, M.; Ghassemi, M.; Moody, B.; Szolovits, P.; Anthony~Celi, L.; and Mark, R.~G. 2016.
\newblock {MIMIC}-{III}, a freely accessible critical care database.
\newblock \emph{Scientific Data}, 3(1): 160035.
\newblock Publisher: Nature Publishing Group.

\bibitem[{{KDIGO}(2012)}]{kdigo_section_2012}
{KDIGO}. 2012.
\newblock Section 2: {AKI} {Definition}.
\newblock \emph{Kidney International Supplements}, 2(1): 19--36.

\bibitem[{Kong, Lin, and Hu(2020)}]{kong_using_2020}
Kong, G.; Lin, K.; and Hu, Y. 2020.
\newblock Using machine learning methods to predict in-hospital mortality of sepsis patients in the {ICU}.
\newblock \emph{BMC Medical Informatics and Decision Making}, 20(1): 251.

\bibitem[{Labach et~al.(2023)Labach, Pokhrel, Huang, Zuberi, Yi, Volkovs, Poutanen, and Krishnan}]{labach_duett_2023}
Labach, A.; Pokhrel, A.; Huang, X.~S.; Zuberi, S.; Yi, S.~E.; Volkovs, M.; Poutanen, T.; and Krishnan, R.~G. 2023.
\newblock {DuETT}: {Dual} {Event} {Time} {Transformer} for {Electronic} {Health} {Records}.
\newblock ArXiv:2304.13017 [cs].

\bibitem[{Lee et~al.(2020)Lee, Yoon, Kim, Kim, Kim, So, and Kang}]{lee_biobert_2020}
Lee, J.; Yoon, W.; Kim, S.; Kim, D.; Kim, S.; So, C.~H.; and Kang, J. 2020.
\newblock {BioBERT}: a pre-trained biomedical language representation model for biomedical text mining.
\newblock \emph{Bioinformatics}, 36(4): 1234--1240.

\bibitem[{Lentzen et~al.(2023)Lentzen, Linden, Veeranki, Madan, Kramer, Leodolter, and Fröhlich}]{lentzen_exmed-bert_2023}
Lentzen, M.; Linden, T.; Veeranki, S.; Madan, S.; Kramer, D.; Leodolter, W.; and Fröhlich, H. 2023.
\newblock {ExMed}-{BERT}: {A} {Transformer}-{Based} {Model} {Trained} on {Large} {Scale} {Claims} {Data} for {Prediction} of {Severe} {COVID}-19 {Disease} {Progression}.
\newblock \emph{IEEE Journal of Biomedical and Health Informatics}, 27(9): 4548--4558.

\bibitem[{Li et~al.(2021)Li, Mamouei, Salimi-Khorshidi, Rao, Hassaine, Canoy, Lukasiewicz, and Rahimi}]{li_hi-behrt_2021}
Li, Y.; Mamouei, M.; Salimi-Khorshidi, G.; Rao, S.; Hassaine, A.; Canoy, D.; Lukasiewicz, T.; and Rahimi, K. 2021.
\newblock Hi-{BEHRT}: {Hierarchical} {Transformer}-based model for accurate prediction of clinical events using multimodal longitudinal electronic health records.
\newblock ArXiv:2106.11360 [cs].

\bibitem[{Li et~al.(2020)Li, Rao, Solares, Hassaine, Ramakrishnan, Canoy, Zhu, Rahimi, and Salimi-Khorshidi}]{li_behrt_2020}
Li, Y.; Rao, S.; Solares, J. R.~A.; Hassaine, A.; Ramakrishnan, R.; Canoy, D.; Zhu, Y.; Rahimi, K.; and Salimi-Khorshidi, G. 2020.
\newblock {BEHRT}: {Transformer} for {Electronic} {Health} {Records}.
\newblock \emph{Scientific Reports}, 10(1): 7155.

\bibitem[{Meng et~al.(2021)Meng, Speier, Ong, and Arnold}]{meng_brltm_2021}
Meng, Y.; Speier, W.; Ong, M.~K.; and Arnold, C.~W. 2021.
\newblock {BRLTM}: {Bidirectional} {Representation} {Learning} {From} {Transformers} {Using} {Multimodal} {Electronic} {Health} {Record} {Data} to {Predict} {Depression}.
\newblock \emph{IEEE Journal of Biomedical and Health Informatics}, 25(8): 3121--3129.

\bibitem[{Pollard et~al.(2018)Pollard, Johnson, Raffa, Celi, Mark, and Badawi}]{pollard_eicu_2018}
Pollard, T.~J.; Johnson, A. E.~W.; Raffa, J.~D.; Celi, L.~A.; Mark, R.~G.; and Badawi, O. 2018.
\newblock The {eICU} {Collaborative} {Research} {Database}, a freely available multi-center database for critical care research.
\newblock \emph{Scientific Data}, 5(1): 180178.

\bibitem[{Poulain and Beheshti(2023)}]{poulain_gt-behrt_2023}
Poulain, R.; and Beheshti, R. 2023.
\newblock {GT}-{BEHRT}: {Graph} {Transformers} on {EHRs}: {Better} {Representation} {Improves} {Downstream} {Performance}.

\bibitem[{Prakash et~al.(2021)Prakash, Chilukuri, Ranade, and Viswanathan}]{prakash_rarebert_2021}
Prakash, P. K.~S.; Chilukuri, S.; Ranade, N.; and Viswanathan, S. 2021.
\newblock {RareBERT}: {Transformer} {Architecture} for {Rare} {Disease} {Patient} {Identification} using {Administrative} {Claims}.
\newblock \emph{Proceedings of the AAAI Conference on Artificial Intelligence}, 35(1): 453--460.
\newblock Number: 1.

\bibitem[{Rasmy et~al.(2021)Rasmy, Xiang, Xie, Tao, and Zhi}]{rasmy_med-bert_2021}
Rasmy, L.; Xiang, Y.; Xie, Z.; Tao, C.; and Zhi, D. 2021.
\newblock Med-{BERT}: pretrained contextualized embeddings on large-scale structured electronic health records for disease prediction.
\newblock \emph{npj Digital Medicine}, 4(1): 86.

\bibitem[{Rupp, Peter, and Pattipaka(2023)}]{rupp_exbehrt_2023}
Rupp, M.; Peter, O.; and Pattipaka, T. 2023.
\newblock {ExBEHRT}: {Extended} {Transformer} for {Electronic} {Health} {Records} to {Predict} {Disease} {Subtypes} \& {Progressions}.
\newblock volume 13932, 73--84.
\newblock ArXiv:2303.12364 [cs].

\bibitem[{Savcisens et~al.(2023)Savcisens, Eliassi-Rad, Hansen, Mortensen, Lilleholt, Rogers, Zettler, and Lehmann}]{savcisens_life2vec_2023}
Savcisens, G.; Eliassi-Rad, T.; Hansen, L.~K.; Mortensen, L.; Lilleholt, L.; Rogers, A.; Zettler, I.; and Lehmann, S. 2023.
\newblock life2vec: {Using} {Sequences} of {Life}-events to {Predict} {Human} {Lives}.
\newblock \emph{Nature Computational Science}.
\newblock ArXiv:2306.03009 [cs, stat].

\bibitem[{Shukla and Marlin(2021)}]{shukla_mtan_2021}
Shukla, S.~N.; and Marlin, B.~M. 2021.
\newblock {mTAN}: {Multi}-{Time} {Attention} {Networks} for {Irregularly} {Sampled} {Time} {Series}.
\newblock ArXiv:2101.10318 [cs].

\bibitem[{Tipirneni and Reddy(2022)}]{tipirneni_strats_2022}
Tipirneni, S.; and Reddy, C.~K. 2022.
\newblock {STraTS}: {Self}-{Supervised} {Transformer} for {Sparse} and {Irregularly} {Sampled} {Multivariate} {Clinical} {Time}-{Series}.
\newblock \emph{ACM Transactions on Knowledge Discovery from Data}, 16(6): 1--17.

\bibitem[{van~de Water et~al.(2024)van~de Water, Schmidt, Elbers, Thoral, Arnrich, and Rockenschaub}]{van_de_water_yet_2024}
van~de Water, R.; Schmidt, H.; Elbers, P.; Thoral, P.; Arnrich, B.; and Rockenschaub, P. 2024.
\newblock Yet {Another} {ICU} {Benchmark}: {A} {Flexible} {Multi}-{Center} {Framework} for {Clinical} {ML}.
\newblock ArXiv:2306.05109 [cs].

\bibitem[{Vaswani et~al.(2017)Vaswani, Shazeer, Parmar, Uszkoreit, Jones, Gomez, Kaiser, and Polosukhin}]{vaswani_attention_2017}
Vaswani, A.; Shazeer, N.; Parmar, N.; Uszkoreit, J.; Jones, L.; Gomez, A.~N.; Kaiser, L.; and Polosukhin, I. 2017.
\newblock Attention {Is} {All} {You} {Need}.
\newblock ArXiv:1706.03762 [cs].

\bibitem[{Wiens and Shenoy(2018)}]{wiens_machine_2018}
Wiens, J.; and Shenoy, E.~S. 2018.
\newblock Machine {Learning} for {Healthcare}: {On} the {Verge} of a {Major} {Shift} in {Healthcare} {Epidemiology}.
\newblock \emph{Clinical Infectious Diseases}, 66(1): 149--153.

\bibitem[{Wolf et~al.(2020)Wolf, Debut, Sanh, Chaumond, Delangue, Moi, Cistac, Rault, Louf, Funtowicz, Davison, Shleifer, von Platen, Ma, Jernite, Plu, Xu, Scao, Gugger, Drame, Lhoest, and Rush}]{wolf_huggingfaces_2020}
Wolf, T.; Debut, L.; Sanh, V.; Chaumond, J.; Delangue, C.; Moi, A.; Cistac, P.; Rault, T.; Louf, R.; Funtowicz, M.; Davison, J.; Shleifer, S.; von Platen, P.; Ma, C.; Jernite, Y.; Plu, J.; Xu, C.; Scao, T.~L.; Gugger, S.; Drame, M.; Lhoest, Q.; and Rush, A.~M. 2020.
\newblock {HuggingFace}'s {Transformers}: {State}-of-the-art {Natural} {Language} {Processing}.
\newblock ArXiv:1910.03771 [cs].

\bibitem[{Wornow et~al.(2023)Wornow, Thapa, Steinberg, Fries, and Shah}]{wornow_ehrshot_2023}
Wornow, M.; Thapa, R.; Steinberg, E.; Fries, J.~A.; and Shah, N.~H. 2023.
\newblock {EHRSHOT}: {An} {EHR} {Benchmark} for {Few}-{Shot} {Evaluation} of {Foundation} {Models}.
\newblock ArXiv:2307.02028 [cs].

\bibitem[{Zhang et~al.(2022)Zhang, Zeman, Tsiligkaridis, and Zitnik}]{zhang_graph-guided_2022}
Zhang, X.; Zeman, M.; Tsiligkaridis, T.; and Zitnik, M. 2022.
\newblock Graph-{Guided} {Network} for {Irregularly} {Sampled} {Multivariate} {Time} {Series}.
\newblock ArXiv:2110.05357 [cs].

\end{thebibliography}

\end{document}